\pdfoutput=1

\documentclass[11pt]{article}

\usepackage{authblk}
\usepackage[]{ACL2023}
\usepackage[none]{hyphenat}

\usepackage{times}
\usepackage{latexsym}
\usepackage{graphicx}
\usepackage{amssymb}

\usepackage{physics,amsmath}
\usepackage{cleveref}
\usepackage{enumitem}
\usepackage{booktabs}

\usepackage{textcomp, xspace}
\usepackage{pbox}
\usepackage{array}
\usepackage{multirow}
\usepackage{xcolor}
\usepackage[normalem]{ulem}
\usepackage{soul}
\crefformat{section}{\S#2#1#3} 
\crefformat{subsection}{\S#2#1#3}
\crefformat{subsubsection}{\S#2#1#3}
\usepackage{arydshln}

\usepackage[T1]{fontenc}

\usepackage[utf8]{inputenc}

\usepackage{microtype}

\usepackage{inconsolata}

\newcommand{\model}{\texttt{ULTRA}\xspace}
\newcommand{\modelENSEMBLE}{\texttt{ULTRA}+\xspace}
\newcommand{\leafer}{\texttt{LEAFER}\xspace}

\newif\ifprintcomments

\newcommand{\change}[1]{\textcolor{black}{#1}}

\newcommand{\red}[1]{\colorbox{gray!15}{#1}}

\title{ULTRA: Unleash LLMs' Potential for Event Argument Extraction \\ through Hierarchical Modeling and Pair-wise Self-Refinement}

\author[1,\thanks{\quad Work done during XFZ's internship at Bloomberg AI.}]{\textbf{Xinliang Frederick Zhang}}
\author[2]{\textbf{Carter Blum}}
\author[2]{\\\textbf{Temma Choji}}
\author[2]{\textbf{Shalin Shah}}
\author[2]{\textbf{Alakananda Vempala}}

\affil[1]{Computer Science and Engineering, University of Michigan}
\affil[2]{Bloomberg}

\affil[1]{\texttt{xlfzhang@umich.edu}}
\affil[2]{\texttt{\{cblum18, tchoji, sshah804, avempala\}@bloomberg.net}}

\begin{document}
\printcommentsfalse
\maketitle
\vspace{-10mm}
\begin{abstract}
Structural extraction of events within discourse is critical since it avails a deeper understanding of communication patterns and behavior trends.
Event argument extraction (EAE), at the core of event-centric understanding, is the task of identifying role-specific text spans (i.e., \textit{arguments}) for a given event. Document-level EAE (DocEAE) focuses on arguments that are scattered across an entire document.
In this work, we explore open-source Large Language Models (LLMs) for DocEAE, and propose \model, a hierarchical framework that extracts event arguments more cost-effectively. Further, it alleviates the \textit{positional bias} issue intrinsic to LLMs. 
\model sequentially reads text chunks of a document to generate a candidate argument set, upon which non-pertinent candidates are dropped through self-refinement. 
We introduce \leafer to address the challenge LLMs face in locating the exact boundary of an argument. \model outperforms strong baselines, including strong supervised models and ChatGPT, by $9.8\%$ when evaluated by Exact Match (EM).

\end{abstract}
\section{Introduction}
\begin{table}[t]
\centering
\resizebox{0.9\linewidth}{!}{%
\begin{tabular}{p{1.5\linewidth}l} \toprule
\textbf{News title:} Drought puts 2.1 million Kenyans at risk of starvation                                                                                                                                                                                                                                                                                                                                                                                                                                                                                                                      \\
\textbf{News body:} \\ 
{[}0{]} National disaster declared as crops fail after poor rains and locusts, while ethnic conflicts add to crisis Last modified on Wed 15 Sep 2021 07.02 BST. \\
{[}1{]} An estimated 2.1 million Kenyans face starvation due to a drought in half the country, which is affecting harvests. \\
{[}2{]} The National Drought Management Authority (NDMA) said people living in 23 counties across the arid north, northeastern and coastal parts of the country will be in “urgent need” of food aid over the next six months, after poor rains \red{between March and May this year}.  \\ 
{[}3{]} The crisis has been compounded by Covid-19 and previous poor rains, it said, predicting the situation will get worse by the end of the year, as October to December rains are expected to be below normal levels.  \\ $\cdot\cdot\cdot$ \\
{[}6{]} In July, the UN Food and Agriculture Organization in Kenya said the country needed 9.4bn Kenyan shillings (£62m) to mitigate the effects of the drought between July and November.  $\cdot\cdot\cdot$ \\
\textbf{Event type:} Droughts \;\;\;\;\;\;\;\;\;\;\;\; \textbf{Argument role:} Date                                                                                                                                                                                                                                                                                                                                                                                                                                                                                                                                                \\ \midrule                                                                        
\textbf{Baseline model outputs:}                                                                                                                                                                                                                                                                                                                                                                                                                                                                                                                                                                           \\ \hdashline[5pt/4pt]
\textbf{Flan-UL2:} Wed 15 Sep 2021        \;\;\;\;\;\;\;\;\;\; \textbf{ChatGPT:} Wed 15 Sep 2021                                                                                                                                                                                                                                                                                                                                                                                                                                                                                                                                                                \\ \midrule
\textbf{\model outputs}                                                                                                                                                                                                                                                                                                                                                                                                                                                \\ \hdashline[5pt/4pt]
\textbf{Layer-1 only:} \{March and May, July, Wed 15 Sep 2021\}                                                                                                                                                                                                                                                                                                                                                                                                                                                                                                                            \\
\textbf{Layer-1 + \leafer:} \{\red{between March and May this year}, July, Wed 15 Sep 2021\}                                                                                                                                                                                                                                                                                                                                                                                                                                                                                                             \\
\textbf{Full model:} \{\red{between March and May this year}, Wed 15 Sep 2021\} \\  \bottomrule                                                                                                                 
\end{tabular}
}
\vspace{-1mm}
\caption{Sample example from DocEE dataset, and outputs of select baselines and \model. The ground-truth span is \red{between March and May this year}. \model can correct itself with the \leafer module, and drops less-pertinent candidates like ``July''. In contrast, both Flan-UL2 and ChatGPT fail to extract since sentence {[}0{]} includes a strong distractor, \textit{``Wed 15 Sep 2021''}.
}
\vspace{-6mm}
\label{tbl:example}
\end{table}
\vspace{-2mm}
Event extraction, a long-standing and prominent information extraction task, aims to extract event structures consisting of core information elements (e.g., ``who'' did ``what'' to ``whom'', ``when'', ``where'', and ``why'') from unstructured texts \citep{Mourelatos1978EventsPA, Riloff1996AutomaticallyGE, ace2005}. Event-centric \hbox{understanding} is of great importance, not only in its inherent merits, but also due to its role as an information-rich representation for downstream tasks like summarization \citep{Marujo2017EventbasedSU, li-etal-2021-timeline}, recommendation \citep{lu-etal-2016-cross, li-etal-2020-gaia}, and news narrative understanding \citep{jin-etal-2022-event, zhang-etal-2022-generative, 10.1145/3584741}. Event argument extraction (EAE), a crucial and challenging step in Event Extraction,
is the task of identifying role-specific text spans (i.e., \textit{arguments}) for a given event \citep{nguyen-etal-2016-joint-event, kar-etal-2020-event}.

\begin{figure*}[t]
    \centering
    \vspace{-10mm}
    \includegraphics[width=0.7\textwidth]{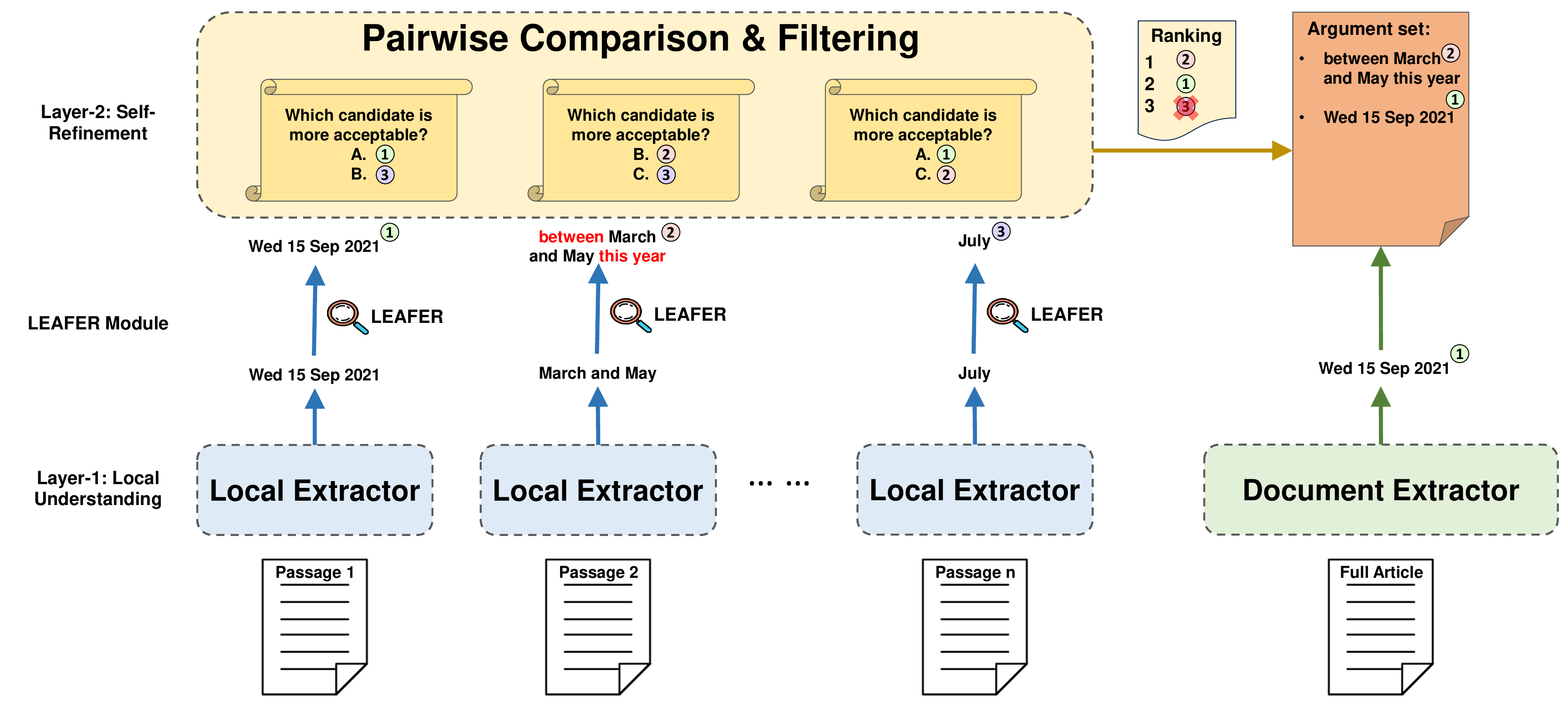}
    \vspace{-2mm}
    \caption{ The overall architecture of \modelENSEMBLE, which consists of \model (\textit{left part}) and a document extractor (\textit{bottom right}). 
    In \model, local extractors (layer-1) first generate a candidate argument set by comprehending text chunks sequentially, upon which \textit{self-refinement} (layer-2) is performed through pairwise comparison to filter out less pertinent candidates. The predicted boundaries in the initial candidate set are rectified by the \leafer module. 
    }
    \label{fig:architecture}
    \vspace{-4mm}
\end{figure*}

Existing EAE research mainly focuses on sentence-level understanding \citep{chen-etal-2015-event, du-cardie-2020-event, lu-etal-2021-text2event} on the \hbox{prevalent} ACE 2005 dataset \citep{ace2005}. Yet, in news, events are usually described at the document level, and arguments are typically scattered across an entire article \citep{Hamborg2019Giveme5W1HAU}. Thus, there is a \hbox{pressing} need to systematically study the document-level EAE (DocEAE) task, since sentence-level EAE systems fail to accommodate long-distance \hbox{dependency} \citep{ebner-etal-2020-multi}, cross-sentence inference \citep{li-etal-2021-document} and multi-answer \citep{tong-etal-2022-docee} problems intrinsic to DocEAE. 
Traditional \hbox{supervised} \hbox{approaches} consume large-scale \hbox{annotations} \citep[e.g.,][more than 30,000 \hbox{annotated} articles required]{zheng-etal-2019-doc2edag, pouran-ben-veyseh-etal-2022-mee} in order to excel, and the state-of-the-art  EAE model requires manual designs of templates for each argument role \citep{hsu-etal-2022-degree}. These approaches are not only costly but also not generalizable, since they cannot handle emerging events \citep{yang-etal-2023-shot}.\footnote{COVID-19 became an emerging topic since 2020 \citep{wang-etal-2020-cord, zhang-etal-2021-cough}, but not covered in traditional EE corpora \citep{ace2005, ebner-etal-2020-multi}.} 
Recently, there has been a notable surge in applications of Large Language Models (LLMs) for NLP tasks, especially closed
models such as Claude \citep{Bai2022ConstitutionalAH}, PaLM \citep{Chowdhery2022PaLMSL}  and GPT-4 \citep{OpenAI2023GPT4TR}. The most relevant works to ours are \citet{Li2023EvaluatingCI, Han2023IsIE}, but they only perform preliminary analysis by assessing ChatGPT's capability of solving IE tasks. \hbox{Meanwhile}, there is no prior research that has attempted to leverage LLMs to tackle DocEAE. In our preliminary investigation, we identified at least three challenges that arise when employing closed LLMs: \change{1) hitting endpoints incurs substantial costs and poses scalability challenges at inference;} 
2) undesirable prompt hacking is needed to ensure performance \citep{Ouyang2022TrainingLM}; 3) given the nature of news, \change{where information is spread across the article}, 
LLMs suffer from the \textit{positional bias issue} \citep[a.k.a, \textit{lost in the middle;}][]{Hou2023LargeLM, Liu2023LostIT}. 
Detailed literature review is in \Cref{sec:related_work}.

To this end, we propose an easy-to-use framework that \textbf{U}nleashes \textbf{L}LMs' potential for event argument ex\textbf{TRA}ction through hierarchical modeling and pair-wise refinement, dubbed \model. \model, built on Flan-UL2 \citep{Tay2022UL2UL},
first sequentially reads text chunks of a news article to generate a candidate argument set. \model then learns to drop less-pertinent candidates through \textit{self-refinement} by means of pairwise comparison. The \leafer module, \textbf{LEA}rning \textbf{F}rom \textbf{ER}rors, is implemented to improve boundary identification of an argument span. Finally, we augment \model with a document-level extractor to capture arguments that require reasoning 
of the full article.

Results on DocEE benchmark \citep{tong-etal-2022-docee} show that \model outperforms strong baselines, e.g., previous state-of-the-art supervised models and ChatGPT, by at least $9.8\%$ and $7.5\%$ when evaluated by the Exact Match (EM) and Head Noun Phrase Match (HM) metrics, but at a considerably reduced monetary cost. Existing methods only cater to improving precision, while our \model significantly boosts recall as well ($39.4$ EM vs. $25.2$). 
Besides better performance and lower costs, \model also doesn't require specialized prompts, 
alleviates the \textit{positional bias} issue, and grants stronger generalizability.

\vspace{-2mm}
\section{Methodology}
\label{sec:method}
\vspace{-2mm}
Taking as input a news article $\mathbf{d}$, \model first reads text chunks of the article $\mathbf{d}$ sequentially to generate a candidate argument set $\{\mathbf{a}\}$ (\Cref{sec:layer_1}), upon which \model drops less-pertinent candidates through self-refinement and returns $\{\mathbf{a}^f\}$ (\cref{sec:layer_2}). A  \leafer module, \textbf{LEA}rning \textbf{F}rom \textbf{ER}rors (\cref{sec:leafer}), is introduced to tackle LLMs' incapability of locating exact boundaries of argument spans, and yield $\{\mathbf{a'}\}$. \modelENSEMBLE 
is a variant augmented with extractions by a document-level extractor to capture information that requires full-article discourse analysis (e.g., extracting ``why''-type arguments; \Cref{sec:ensemble}).  Figure~\ref{fig:architecture} depicts the overall framework of \modelENSEMBLE.  

Putting it all together, we produce two versions: \model-base and \model-long, which consume 5-sentence and 15-sentence windows in layer-1, respectively. \change{Instead of conducting costly prompt hacking \citep{Ouyang2022TrainingLM}, we adopt an existing instruction from NIv2 \citep{wang-etal-2022-super} and tailor it to our use case, named \textbf{aligned instruction}. We show designed task instructions  $\{\mathbf{I}\}$ in Table~\ref{tbl:prompts}. } 

\subsection{Layer-1: Local Understanding}
\label{sec:layer_1}
Given a document $\mathbf{d}$, we first divide  $\mathbf{d}$ to multiple $k$-sentence passage windows with a step size of $\lfloor\frac{k}{2}\rfloor$, denoted as $\{w_1, w_2, \cdot\cdot\cdot, w_l\}$. We adopt a fixed-window-size approach instead of a fixed-sequence-length approach \citep{devlin-etal-2019-bert, Sun2019HowTF, Pappagari2019HierarchicalTF}, which might cut a sentence in the middle, to allow each local extractor to comprehend each passage window in its entirety. Instantiated with Flan-UL2, the \textbf{local extractor} takes as input the concatenation of a task instruction ($\mathbf{I}$), a passage window ($w_i$), and a question written in natural language ($q_j$), e.g., \textit{What is the ``date'' for the ``Tsunami'' event?} We prompt the local extractor in a zero-shot fashion\footnote{We observe that few-shot prompting yields inferior results} and explicitly instruct it to generate \textit{N/A} if the input passage does not contain any relevant answer. After deduplication, we end up with a candidate argument set $\{\mathbf{a}\}_j$ for each question $q_j$.\footnote{For brevity, we omit the subscript $j$ in main contents.}

\vspace{-2mm}
\subsection{\leafer Module}
\label{sec:leafer}
\vspace{-1mm}

LLMs are deemed to have a knack for extracting relevant information \citep{Li2023EvaluatingCI, Han2023IsIE}, but we notice that LLMs still suffer from pinpointing the exact boundary of an argument span. Specifically, as shown in Figure~\ref{fig:architecture}, local extractions ($\{\mathbf{a}\}$) contain an apparently sensible answer ``March and May'' to the question ``What is the `Date' for the `Droughts' event?'', which is \textit{lexically similar but semantically different} from the ground-truth answer ``between March and May this year''.  
To this end, we introduce a new module, \leafer, short for \textit{\textbf{LEA}rning \textbf{F}rom \textbf{ER}rors}, to alleviate this issue. The \leafer module is a small-scale LM trained on errors produced by Flan-UL2. The trained \leafer is employed to generate a \textbf{judgment}, which is to explicitly inform what is wrong and why it is wrong. The insightful judgment enables \model to rectify boundaries of candidate arguments in $\{\mathbf{a}\}$ and produce $\{\mathbf{a'}\}$.

To support the training of \leafer, we construct a \leafer Bank using the few-shot training set of 50 annotated articles. Specifically, we prompt the same layer-1 local extractor to extract arguments for each (text chunk, question) input pair using the approach outlined in \cref{sec:layer_1}. For each input pair, we match the machine-extracted argument span with the corresponding ground-truth answer to produce a \textbf{judgment} automatically.
Then, \leafer is fine-tuned on this   \leafer Bank and trained to generate a judgment given an input pair and the machine-extracted answer. 
In this study, we instantiate the \leafer module with Flan-T5-large.

\vspace{-2mm}
\subsection{Layer-2: Self-Refinement}
\label{sec:layer_2}
\vspace{-2mm}
 While \leafer addresses the semantic drift and imprecise boundary issues, \model exhibits an \textit{over-generation} issue due to window-based local extractors.
Seeing recent success leveraging an LLM as a judge \citep{Zheng2023JudgingLW, Wang2023LargeLM}, we propose a \textbf{self-refinement} module that allows \model to reflect on candidate arguments ($\{\mathbf{a'}\}$), and drop less pertinent candidates through pairwise ranking. There are usually two variations of LLM-as-a-judge: single-answer grading and pairwise comparison. As studied in \citet{Zheng2023JudgingLW} and observed in our preliminary study, we find that single answer grading cannot serve as an effective refinement judge since 1) absolute scores are extremely inflated and a considerably large portion of scores are close to 1 on a scale of 0 to 1; and 2) single answer grading fails to capture subtle differences between a specific pair. Therefore, in layer-2, we leverage \textbf{ranking by pairwise comparison} \citep{Jamieson2011ActiveRU, lee-vajjala-2022-neural, jiang-etal-2023-llm} to obtain the final argument set, $\{\mathbf{a}^f\}$, by first prompting Flan-UL2 to pick a better answer between a candidate pair, then ranking all candidates by aggregating pairwise-comparison scores, and finally filtering out candidates at low positions. To support dynamic filtering, we decide on $|\{\mathbf{a}^f\}|$ as follows: 

\vspace{-4mm}
\begin{equation}
  |\{\mathbf{a}^f\}| =\lfloor 1+\log_2(|\{\mathbf{a'}\}|)\rfloor 
\end{equation}

The pairwise comparison produces a non-trivial score 
and catches nuanced differences, but is still trapped by the \textit{positional bias} \citep{ko-etal-2020-look, Wang2023LargeLM, Liu2023LostIT} and \textit{lack of scalability} \change{due to the quadratic growth in pairwise comparisons}.
To mitigate these two issues, we resort to calibration and pruning, respectively.

\vspace{-2mm}
\paragraph{Calibration.} In layer-2, positional bias refers to a model's tendency to assign a higher score to an option at a particular position in a list, which has been shown to exist in ChatGPT/GPT-4 \citep{Wang2023LargeLM}. The issue is manifested as Flan-UL2 biasing towards an earlier displayed candidate. 
Drawing on the \textit{Contextual Calibration} \citep{calibration}, as demonstrated in \cref{eq:calibration}, we calibrate the raw probabilities of each option between a pair to reveal the truthful probabilities, i.e., $\textrm{P}(\mathbf{a_i} | \mathbf{d})$.

\vspace{-6mm}
\begin{equation}
  \textrm{P}(\mathbf{a_i} | \mathbf{d}) = \textrm{softmax}(\mathbf{g(\textrm{P}(\mathbf{a_i} | \mathbf{d}, \mathbf{I}; \theta), \textrm{P}(\mathbf{a_i} | \mathbf{I}; \theta))})
\label{eq:calibration}
\end{equation}

\noindent where $\textrm{P}(\mathbf{a_i}|\cdot)$ denotes the  probability of an argument $\mathbf{a_i}$ being preferred given a certain input, and $\mathbf{d}$ and $\mathbf{I}$ denote the article and task instruction (see Table~\ref{tbl:prompts} for the instruction). Following \citet{calibration}, $\mathbf{g(x,y)}$ is a calibration function that can be instantiated as additive, $\mathbf{g(x,y)} = \mathbf{x} - \mathbf{y}$, or multiplicative functions, $\mathbf{g(x,y)} = \frac{\mathbf{x}}{\mathbf{y}}$.
Using our designed comparison instruction ($\mathbf{I}$), we compute the prior probability $\textrm{P}(\mathbf{a_i} |  \mathbf{I}; \theta)$  by leaving the \textit{$\{$article$\}$} field blank, while we fill in a concrete article when computing raw probability $\textrm{P}(\mathbf{a_i} |\mathbf{d}, \mathbf{I}; \theta)$.
With this calibration, we manage to alleviate the \textit{positional bias} induced by the input template $\mathbf{I}$ and the innate bias of LLMs, $\theta$.

\vspace{-2mm}
\paragraph{Pruning.} To tackle the scalability issue in which the number of comparisons grows quadratically, we prune the candidate set $\{\mathbf{a'}\}$ upfront to shrink its size. Specifically, we design a strategy that aligns with the fundamental principles of news journalism, wherein journalists prioritize the presentation of crucial information at the outset of a news story, known as the ``inverted pyramid'' structure \citep{po2003news, Hamborg2019Giveme5W1HAU, liu-etal-2022-politics}. That is, we consider up to 5 \textit{earliest} candidate arguments, where the earliness of an argument is determined by its first occurrence in a news article.
Our pruning strategy empirically reduces the number of subsequent pairwise computations by half. We also find that pruning itself can improve precision, even without pairwise comparisons. This further illuminates the validity of our designed pruning strategy.

\subsection{Ensembling: \modelENSEMBLE}
\label{sec:ensemble}
 The ensembling technique consistently improves performance for a wide array of NLP tasks \citep{ superglue, ensemble, Pitis2023BoostedPE, jiang-etal-2023-llm}. 
LLM-Blender attempts to ensemble various LLMs on output space \citep{jiang-etal-2023-llm}, which prohibitively demands many computational resources.
 \change{Instead, we suggest a simpler and more efficient approach: merging outputs by both \model and a document-level argument extractor, which reads the full article and a question when extracting arguments. This way, we manage to combine the benefits of both local (\textit{high recall}) and document-level (\textit{high precision}) extractions.}

Similar to \citet{Labrak2023AZA, Han2023IsIE}, we also observe marginal improvement on the dev set when providing in-context examples. To reduce inference-time overhead, we prompt the document-level extractor in a zero-shot manner. 
\begin{table*}[t]
\centering
\vspace{-4mm}
\resizebox{0.8\linewidth}{!}{%
\begin{tabular}{llrrrrrrrr}
\toprule

\multicolumn{1}{c}{\multirow{3}{*}{Category}} & \multicolumn{1}{c}{\multirow{3}{*}{Method}}     & \multicolumn{6}{c}{Performance}                                                                                                                 & \multicolumn{2}{c}{Cost}                               \\ \cmidrule(lr){3-8}    \cmidrule(lr){9-10} 
\multicolumn{1}{c}{}                          & \multicolumn{1}{c}{}                                          & \multicolumn{3}{c}{EM}                                                 & \multicolumn{3}{c}{HM}                                                 & \multirow{2}{*}{Training} & \multirow{2}{*}{Inference} \\ \cmidrule(lr){3-5} \cmidrule(lr){6-8}
\multicolumn{1}{c}{}                          & \multicolumn{1}{c}{}                                          & P & R & F1 & P & R & F1 &                           &                            \\ \midrule
\multirow{2}{*}{Supervised ML}                & EEQA* \citep{du-cardie-2020-event}          & 29.4                  & 20.3                  & 24.0                   & 68.1                  & 46.9                  & 55.5                   & \multirow{2}{*}{\$\$\$}      & \multirow{2}{*}{$\sim$0}   \\
                                              & Onology QA* \citep{tong-etal-2022-docee}     & \textbf{36.6}                  & 25.2                  & 29.8                   & 69.7                  & 48.0                  & 56.9                   &                           &                            \\ \midrule
\multirow{3}{*}{Closed LLM}                & ChatGPT \citep{Li2023EvaluatingCI}         & 35.6                  & 18.0                  & 23.9                   & 74.4                  & 58.0                  & 65.2                   & \multirow{3}{*}{0}  & \multirow{3}{*}{\$-\$\$}        \\
                                              & ChatGPT (single question)                & 30.9                  & 22.7                  & 26.2                   & 63.5                  & 65.3                  & 64.4                   &                           &                            \\
                                              & CoT-ChatGPT \citep{wang-etal-2023-element} & 31.2                  & 16.2                  & 21.3                   & 71.0                  & 55.2                  & 62.1                   &                           &                            \\ \midrule
\multirow{2}{*}{Flan-UL2}                     & Custom instructions**                                           & 27.6                  & 17.8                  & 21.6                   & 69.2                  & 45.2                  & 54.6                   & \multirow{2}{*}{\$}  & \multirow{2}{*}{$\sim$0}        \\
                                              & Aligned instruction                                  & 36.1                  & 20.7                  & 26.3                   & \textbf{76.6}         & 52.0                  & 62.0                   &                           &                            \\
                                               \midrule
\multirow{4}{*}{\model (Ours)} & \model-base                                 & 29.0                  & 34.5                  & 31.5                   & 61.8                  & 70.3                  & 65.8                   & \multirow{4}{*}{\$}  & \multirow{4}{*}{$\sim$0}        \\
                                              & \; + Ensemble (i.e., \modelENSEMBLE)                                & 28.0                  & \textbf{39.4}         & \textbf{32.7}          & 63.7                  & \textbf{75.3}         & 69.0                   &                           &                            \\
                                              & \model-long                                  & 32.3                  & 30.5                  & 31.4                   & 68.4                  & 65.9                  & 67.1                   &                           &                            \\
                                              & \; + Ensemble (i.e., \modelENSEMBLE)                                     & 30.2                  & 35.5                  & 32.6          & 68.6                  & 71.5                  & \textbf{70.1}          &                           &                           \\ \bottomrule
\end{tabular}
}
\caption{Results on DocEE dataset for document-level event argument extraction, and breakdown of EM and HM scores by precision (P), recall (R) and F1. We also report estimated monetary cost by model category, divided into training and inference costs (\Cref{sec:cost_estimate}). Best results are \textbf{bold}. \model achieves the best F1 performances at a reduced cost, and reaches higher recalls than any baseline.  
We use the additive function when performing calibration (\Cref{sec:layer_2}), though we see a trivial difference between additive and multiplicative functions.
The ablation study results of \model can be found in Table~\ref{tbl:results_additional}. *Results are taken from \citet{tong-etal-2022-docee}. **Average results of 5 instructions, results of individual custom instructions are included in Table~\ref{tbl:custom_results}. 
} 
\vspace{-5mm}
\label{tbl:results}
\end{table*}

\vspace{-2mm}
\section{Experiments}
\label{sec:experiments}
\vspace{-2mm}
We conduct experiments on the DocEE dataset \citep{tong-etal-2022-docee}, which contains 27,485 news articles, classified into 59 event types, and 356 argument roles. We use the cross-domain setting in our experiments since it only contains a minimally annotated target training set (i.e., 50 articles) which can best assess models' generalizability in the wild. Specifically, its test set contains $1,955$ news articles covering $10$ different event types, and each article is annotated with $\sim6.5$ arguments. 
We use the same data split and processed texts as in the original DocEAE dataset for a fair comparison.

In terms of evaluation metrics, we follow the literature on document-level event argument extraction \citep{du-cardie-2020-document, tong-etal-2022-docee}, and adopt Exact Match (EM) and Head Noun Phrase Match (HM) as evaluation metrics. EM assesses if an extracted argument exactly matches a reference, while HM is a relaxed metric that concerns if there is an overlap of head words of noun phrases between extractions and references.

\vspace{-2mm}
\paragraph{Baselines.}
We compare \model against three model families to comprehensively study extraction performance and monetary costs.
The first model family, \textit{Supervised ML}, characterized as using human annotations as the supervision signal to train small-scale LMs, consists of EEQA \citep{du-cardie-2020-event} and Ontology  QA \citep{tong-etal-2022-docee}. Ontology QA, an extension of EEQA, incorporates argument ontology knowledge, which achieves the SOTA performance for DocEAE. 
Second, we compare with ChatGPT using different prompting techniques, given its popularity and impressive capability. We follow \citet{Li2023EvaluatingCI} and prompt ChatGPT to extract spans for \textit{all argument roles} in one pass. For single-question variant, we modify the original prompt to instruct ChatGPT to extract span(s) for \textit{only one argument role} at a time. Motivated by \citet{wang-etal-2023-element}, which generates a chain-of-thought rationale before summarizing an article, we build the CoT-ChatGPT variant by replacing the summarizer in \citet{wang-etal-2023-element} with an argument extractor.
The last family involves prompting a document-level extractor with different instructions, utilizing Flan-UL2 as its backbone for a fair comparison. 
This serves two purposes: test sensitivity to different custom instructions that are designed from scratch; and illuminate the effectiveness of  \textit{Aligned Instructions}.

\vspace{-2mm}
\section{Results and Analysis}
\vspace{-2mm}
Our proposed \model achieves the best F1 scores across the board (Table~\ref{tbl:results}), especially compared to two strong baseline families, \textit{Supervised ML} and \textit{Closed LLM}, at a considerably reduced monetary cost in training and inference. \model also significantly improves the EM recall by $56\%$ over the best-performing baseline model ($39.4$ vs. $25.2$), demonstrating robust generalizability considering \model's exposure to at most 5-shots per event type.

Seeing the relatively low EM scores, we explore the errors and include case analyses in \Cref{tbl:case_analysis}. We provide cost estimation in \Cref{sec:cost_estimate}. 

\vspace{-2mm}
\paragraph{Using ChatGPT for DocEAE.} Despite the common flaw of outputting seemingly coherent assertions that are false in reality, known as hallucination \citep{Manakul2023SelfCheckGPTZB, Feldman2023TrappingLH}, we recognize another issue, which seems to be less studied in the NLP community, that answer spans extracted by ChatGPT are \textit{verbose} \citep{Zheng2023JudgingLW, Chen2023HowIC}. This explains the reason why ChatGPT achieves the best HM scores in the literature since longer generations are more likely to contain relevant information, while the EM is low due to the \textit{nature of verbosity}.

\paragraph{Further Study on Window Size.}
Despite \model-base and \model-long achieving almost identical EM F1 scores, they present different extraction properties, wherein \model-base reaches the highest recall while \model-long is more balanced. In this subsection, we specifically look into the extraction property of the Layer-1-only variant of \model. Figure~\ref{fig:impact_of_window_size} shows the performance trend with the window size. We notice that precision steadily goes up while recall consistently goes down by increasing the window size. We attribute this trend to the fact that a larger window size leads to fewer text chunks being fed into \model. It is also worth mentioning that the overall F1 performance plateaus after a window size of 15. This observation underscores a key aspect of \model: its flexibility for accommodating various extraction criteria. For instance, when the objective is to harvest the most relevant information, opting for a smaller window size appears to be a favorable choice. Conversely, selecting a larger window is advisable if precision is the core of a product or the target audience consists of vulnerable populations susceptible to misinformation.

\vspace{-2mm}
\section{Conclusion}
\vspace{-2mm}
In this study, we present \model, a cost-effective event argument extraction framework built upon an open-source LLM. Concretely, \model reads a sequence of text chunks from an article, the outputs of which are refined through self-refinement. With minimal annotation efforts, a \leafer module is implemented to improve argument span boundary identification. Our results show the superiority of \model in comparison to \textit{supervised ML} and \textit{closed LLMs}. We further showcase the customizability of \model to cater to different extraction criteria. 
\section*{Limitations}
\paragraph{GPU resources.}
Despite \model managing to reduce monetary cost, it still requires advanced computational resources. Specifically, we deploy \model on a single NVIDIA A100 (80GB) with significant CPU and memory resources. Due to budget constraints, we truncate an input if it is longer than $2,048$ tokens.

\paragraph{DocEE benchmark.}
Due to the limited large-scale, well-regarded datasets for the DocEAE task, we only conduct experiments on one benchmark -- DocEE dataset \citep{tong-etal-2022-docee}. DocEE, though covering $59$ event types and $356$ argument roles, is still not comprehensive. Therefore, our results might not truthfully reveal the generalizability of \model. In future work, we will explore how to examine the true generalizability of developed systems in the wild.

\section*{Acknowledgements}
The authors thank the Bloomberg AI 
team for their support of this work. We would like to thank Shijie Wu for hardware support and discussions on FLAN-UL2 setup configurations. We appreciate Igor Malioutov, Daniel Preo\c{t}iuc-Pietro and Umut Topkara for their insightful inputs and helpful feedback at different stages of this project. We also thank ACL ARR reviewers for their helpful feedback.


\clearpage
\setcounter{table}{0}
\setcounter{figure}{0}
\renewcommand{\thefigure}{A\arabic{figure}}
\renewcommand{\thetable}{A\arabic{table}}

\appendix
\section{Related Work}
\label{sec:related_work}
\subsection{Event Argument Extraction (EAE)}

Most event argument extraction research has experimented on the 2005 Automatic Content Extraction \citep[ACE 2005;][]{ace2005}, while recent work delves into domain-specific areas such as biomedical texts \citep{Zhao2020ANJ,He2022ABE}, legal documents \citep{Li2020EventEF, shen-etal-2020-hierarchical}, partisan contents \citep{liu-etal-2023-things}, morality-bearing contents \citep{zhang2023moka}, and conversations  \citep{Srivastava2023MAILEXEE}.

Existing work primarily focused on the sentence-level event understanding task.  Methods can be categorized under one of the three following approaches: 
sequence labeling \citep{chen-etal-2015-event, nguyen-etal-2016-joint-event} where \citet{lin-etal-2020-joint} further constrains the inference with global features;
question answering \citep{du-cardie-2020-event}, which includes ontology knowledge about argument roles;
and generative information extraction \citep{tanl, lu-etal-2021-text2event}.  \change{Particularly, DEGREE reformulates EAE as template-based conditional generation, and archives impressive performance on various benchmarks \citep{hsu-etal-2022-degree}.} 
Yet, it demands huge annotation efforts, which require one template for each argument role, and is therefore not generalizable. In this work, we are seeking to improve EAE performance with general instructions instead of argument-specific templates. 

Lately, there has been an increasing interest in document-level EAE (DocEAE), since events are usually described at the document level and arguments are usually scattered across multiple sentences \citep{sundheim-1992-overview, Hamborg2019Giveme5W1HAU, tong-etal-2022-docee}. 
For example, RAMS \citep{ebner-etal-2020-multi} and MEE \citep{pouran-ben-veyseh-etal-2022-mee} both define ``document'' as a 5-sentence segment. In contrast, WikiEvents \citep{li-etal-2021-document} and DocEE \citep{tong-etal-2022-docee} present full articles and focus on argument extractions for the main event. In this work, we use DocEE as a benchmark since it features broad coverage of event types in the news domain. 
\change{Methodology-wise, \citet{du-cardie-2020-document} and \citet{tong-etal-2022-docee} handle DocEAE by extending sentence-level labeling and question-answering approaches, respectively. \citet{li-etal-2021-document} frames DocEAE as template-based conditional generation in the same vein as the sentence-level generative approach. 
Unfortunately, none of the aforementioned methods tackle the \textit{argument-scattering} challenge; instead, they treat a full article as if it were an extended sentence. \citet{zheng-etal-2019-doc2edag} is the first work to address this issue by modeling DocEAE as an entity-centric graph, which is further augmented with a ``tracker'' module to capture the interdependency among arguments and events \citep{xu-etal-2021-document}.}
Nonetheless, the ``tracker'' is insufficient due to its limitation of not considering the results of later extractions when processing earlier ones. On the contrary, our \model bridges the gap through the implementation of a \textit{self-refinement} module, which is grounded in pairwise comparison and functions akin to a bi-directional tracker.

\subsection{Using Large Language Models for IE}
The past years have witnessed the rise of transformer architecture \citep{Vaswani2017AttentionIA}, paving the way for a series of powerful language models.
Recent studies have evinced that scaling up model sizes yields more powerful abilities \citep{Hoffmann2022TrainingCL}, and unlocks an emergent ability that is not present in smaller models \citep{Wei2022EmergentAO}. These large language models (LLMs), which often exceed a hundred billion parameters, are typically closed systems (i.e., no open checkpoints available). Notable examples include PaLM \citep{Chowdhery2022PaLMSL}, Claude \citep{Bai2022ConstitutionalAH}, and GPT-4 \citep{OpenAI2023GPT4TR}. Numerous methods are also developed to enhance LLMs' reasoning and problem-solving capabilities, such as chain-of-thought \citep{Wei2022ChainOT}, self-correction \citep{Pan2023AutomaticallyCL}, and external tool (e.g., Python interpreter) augmentation \citep{gao2022pal, chen2022program} among others.

ChatGPT,\footnote{\url{https://chat.openai.com/}} one of the most burgeoning LLM, is trained on high-quality conversation datasets using reinforcement learning from human feedback \citep[RLHF;][]{NIPS2017_d5e2c0ad}, has led to a transformative wave. The most relevant to our research is leveraging ChatGPT for the information extraction task \citep{Li2023EvaluatingCI, Han2023IsIE}, including named entity recognition \citep{Xie2023EmpiricalSO}, temporal relation extraction \citep{yuan-etal-2023-zero}, event detection \citep{Sharif2023CharacterizingIS}, and event argument extraction \citep{Wei2023ZeroShotIE}. These papers' primary focus is either benchmarking ChatGPT's performance, which shows inferior results to specialized supervised IE systems \citep{Li2023EvaluatingCI, Han2023IsIE}, or curating new benchmark datasets \citep{Gao2023BenchmarkingLL}. In contrast, our proposed \model framework outperforms strong baselines, including the previous state-of-the-art (SOTA) models, capitalizing on the effectiveness of our designed \leafer and self-reflection modules. Besides, to the best of our knowledge, we are the first to exploit LLMs for the DocEAE task.

 \begin{figure}[t]
    \centering
    \includegraphics[width=0.5\textwidth]{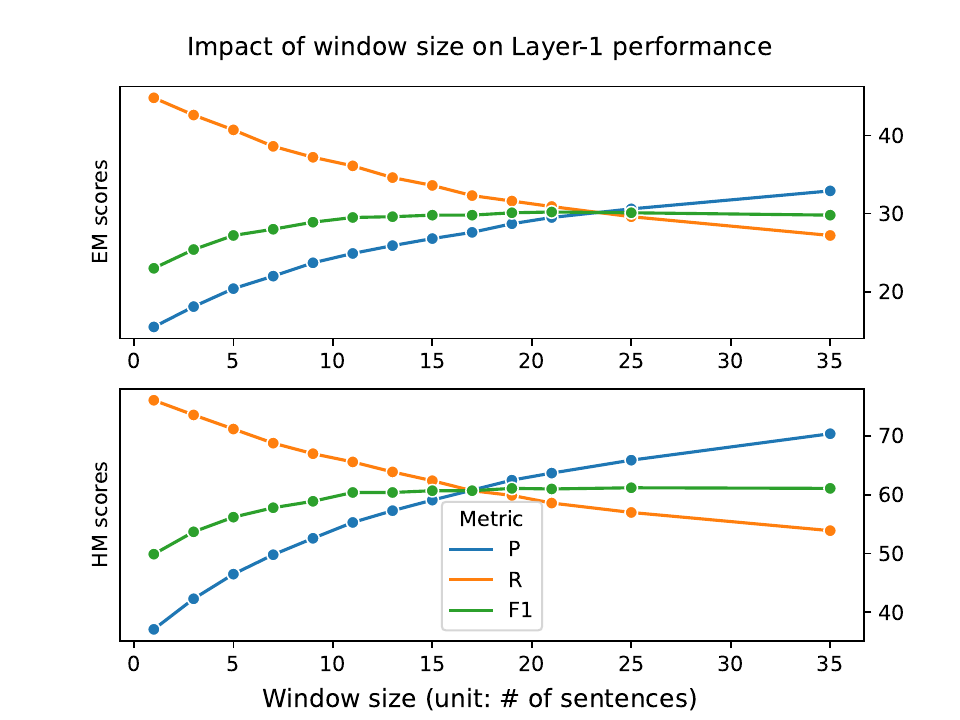}
    \vspace{-6mm}
    \caption{
    The impact of window size on the performance of the Layer-1-only variant of \model. Results are based on dev set. With the window size,  precision goes up while recall goes down since fewer chunks are fed into \model. The F1 performances plateau after the window size of 15. 
    }
    \vspace{-4mm}
    \label{fig:impact_of_window_size}
\end{figure}

 \begin{table}[t]
\centering
\resizebox{1.1\linewidth}{!}{%
\begin{tabular}{p{1.5\linewidth}l} \toprule
\textbf{News title:}     Experts: Oregon seems to be in 'perpetual drought'\\
\textbf{News body:} \\

{[}0{]} Experts say Oregon is becoming less resilient to drought as fewer seasons of abundant rain and snow prevent it from bouncing back from hot and dry conditions. \\

{[}1{]} Wheat at the farm of Nicole Berg in Washington's Horse Heaven Hills shows signs of a drought in May 2021, with a damaged curled head. \\

{[}2{]} Anna King The Capital Press reports that Larry O'Neill, state climatologist at Oregon State University, says the current drought is ``historically significant,'' with about three-quarters of the state experiencing conditions considered ``extreme'' or ``exceptional.'' \\

{[}3{]} However, the state is actually in the fourth year of below-average precipitation, which has exacerbated the drought during ``unprecedentedly'' high temperatures this summer, O'Neill told the Oregon Water Resources Commission on Wednesday. \\

{[}4{]} ``We don't recover from droughts as quickly as we did previously,'' he said. ``We seem to be in perpetual drought.'' \\

{[}5{]} Parched soils were insufficiently recharged with moisture over winter and spring, which has harmed vegetative growth, including crops and forage, said Ryan Andrews, a hydrologist at the Oregon Water Resources Department, which is overseen by the commission. \\

{[}6{]} Reservoir and stream flow levels are below average across most of the state, reducing water available to irrigators, while ranchers have sold off livestock due to poor rangeland conditions, he said. \\

{[}7{]} Fish die-offs followed the June heat wave in several important rivers basins, including the Willamette, Grande Ronde, John Day and along the North coast, Andrews said. \\

{[}8{]} The state would need plentiful rain and snow during the autumn to begin emerging from the drought, but the long-term federal climate forecast doesn't anticipate such a reversal, he said. \\

{[}9{]} ``We're anticipating conditions to persist, at least in the near term.'' Between March and July, the state received less rain than during any comparable period in nearly a century, O'Neill said. \\

{[}10{]} ``The dry spring and summer is one of the main contributing factors to why this drought has become so severe.'' \\

{[}11{]} An IBIS explores what habitat remains on the Klamath Basin's wildlife refuges during a drought year that is exacerbating water resources challenges in this arid region. \\

{[}12{]} Devan Schwartz / OPB The area under ``extreme'' and ``exceptional'' drought ratings is the most extensive in Oregon since the start of the U.S. Drought Monitor more than 20 years ago, he said. \\

{[}13{]} The most severe ``exceptional'' level of drought now seen across one-fourth of the state would normally be expected to occur every 20 to 50 years, O'Neill said. \\

{[}14{]} However, droughts are judged by historical standards, so the concept of such ``recurrence intervals'' grows less valid as dry periods become more common, he said. \\

{[}15{]} ``It's going to take some time to get used to the new normal we're experiencing right now,'' O'Neill said.  \\ \midrule                                                                        
\textbf{Argument role 1:} Related Rivers or Lakes   \\ \hdashline[5pt/4pt]
\textbf{\model:} \{``the Willamette, Grande Ronde, John Day and along the North coast''\}    \\ 
\textbf{Ground-truth:} \{``Willamette, Grande Ronde, John Day and along the North coast''\}                                                                                                          \\ \midrule
\textbf{Argument role 2:} Cause   \\ \hdashline[5pt/4pt]
\textbf{\model:} \{``fewer seasons of abundant rain and snow''\}    \\ 
\textbf{Ground-truth:} \{``The dry spring and summer'', ``fewer seasons of abundant rain and snow''\}      
                                                         \\ \midrule
\textbf{Argument role 3:} Areas Affected   \\ \hdashline[5pt/4pt]
\textbf{\model:} \{``Oregon''\}    \\ 
\textbf{Ground-truth:} \{``Willamette, Grande Ronde, John Day and along the North coast'', ``Washington''\}   \\
\bottomrule                                                                                                                 
\end{tabular}
}
\vspace{-1mm}
\caption{Case analyses of \model outputs for a \texttt{Droughts} event. For argument role 1, the system output is considered an exact match with the ground-truth annotation, since stop words are removed at evaluation time. For argument role 2, our \model achieves 100\% precision but 50\% recall since the model fails to capture both reasons that caused the drought. For argument role 3, the system output is completely wrong.
}
\vspace{-6mm}
\label{tbl:case_analysis}
\end{table}

\begin{table*}[t]
\centering
\resizebox{.65\linewidth}{!}{%
\begin{tabular}{clrrrrrr}
\toprule
\multirow{2}{*}{Model}                      & \multicolumn{1}{c}{\multirow{2}{*}{Configuration}} & \multicolumn{3}{c}{EM}                        & \multicolumn{3}{c}{HM}                        \\ \cmidrule(lr){3-5} \cmidrule(lr){6-8}
                                            & \multicolumn{1}{c}{}                               & P             & R             & F1            & P             & R             & F1            \\ \midrule
\multirow{3}{*}{\model-base} & Layer-1 only                                       & 22.5          & 42.5          & 29.4          & 50.3          & 77.7          & 61.1          \\
                                            & Layer-1 + \leafer                                   & 22.6          & \textbf{43.4} & 29.7          & 50.7          & \textbf{78.4} & 61.6          \\
                                            & Layer-1 + \leafer + Layer-2                         & 29.0          & 34.5          & \textbf{31.5} & 61.8          & 70.3          & 65.8          \\ \midrule
\multirow{3}{*}{\model-long} & Layer-1 only                                       & 29.2          & 34.0          & 31.4          & 63.3          & 68.9          & 66.0          \\
                                            & Layer-1 + \leafer                                   & 29.2          & 34.5          & 31.6          & 63.4          & 69.7          & 66.4          \\
                                            & Layer-1 + \leafer + Layer-2                         & \textbf{32.3} & 30.5          & 31.4          & \textbf{68.4} & 65.9          & \textbf{67.1} \\  \bottomrule
\end{tabular}
}
\caption{Ablation study results of variants of \model. \model-base manages to improve recall by over-generating candidate answers, while the over-generation problem is redressed by self-refinement (layer-2) through pairwise comparison. \model-long does not confront the over-generation issue, thus, layer-2 does not contribute significantly to the performance as in \model-base. Best results are \textbf{bold}. }
\vspace{-2mm}
\label{tbl:results_additional}
\end{table*}

\begin{table*}[t]
\centering
\resizebox{1.0\linewidth}{!}{%
\begin{tabular}{rrrrrrrl}
\toprule \multirow{2}{*}{ID} & \multicolumn{3}{c}{EM} & \multicolumn{3}{c}{HM} & \multicolumn{1}{l}{\multirow{2}{*}{Instruction template}}                                                                                                                                  \\ \cmidrule(lr){2-4}    \cmidrule(lr){5-7}  
                    & P      & R     & F1    & P      & R     & F1    & \multicolumn{1}{r}{}                                                                                                                                                               \\ \midrule
0                   & 26.5   & 18.9  & 22.0  & 67.6   & 45.2  & 54.2  & The following is a news article about a ``\{e\_type\}'':\textbackslash{}n\{news\}\textbackslash{}nBy reading the above article, determine the ``\{arg\_role\}'' for the ``\{e\_type\}''.      \\
1                   & 29.5   & 19.5  & 23.5  & 71.3   & 47.3  & 56.8  & Here is a news article:\textbackslash{}n\{news\}\textbackslash{}nThe above news article is about a ``\{e\_type\}''. Identify ``\{arg\_role\}'' for the ``\{e\_type\}'' from the news article. \\
2                   & 29.0   & 16.2  & 20.8  & 71.8   & 42.8  & 53.7  & Read the following news article, and then answer questions. Context: \{news\} \textbackslash{}nQuestion: Identify ``\{arg\_role\}'' for this ``\{e\_type\}'' event.                         \\
3                   & 25.5   & 18.6  & 21.5  & 64.6   & 43.1  & 51.7  & Given the following news about a ``\{e\_type\}'':\textbackslash{}n\{news\}\textbackslash{}nThe ``\{arg\_role\}'' for the ``\{e\_type\}'' is                                                   \\
4                   & 27.6   & 15.7  & 20.0  & 70.5   & 47.5  & 56.8  & Given the following news about a ``\{e\_type\}'':\textbackslash{}n\{news\}\textbackslash{}nWhat is the ``\{arg\_role\}'' for the ``\{e\_type\}''?   \\ \bottomrule                                         
\end{tabular}
}
\caption{Performances of each individual custom instruction. \{e\_type\}, \{arg\_role\} and \{news\} are placeholders to be filled with event type, argument role and news content, respectively. Flan-UL2 is considerably sensitive to the input instruction, and even with a tiny change in the question, the model performance varies a lot, as manifested by contrasting instruction ID 3 and 4.}
\label{tbl:custom_results}
\vspace{-2mm}
\end{table*}

\begin{table*}[t]
\centering
\resizebox{.95\linewidth}{!}{%
\begin{tabular}{ll}
\toprule
Stage                                    & Instruction                                                                              \\ \midrule
\multirow{3}{*}{Layer-1 local extractor}                    & Given a passage from a news article about \{e\_type\}, select the tokens representing information   \\ & about '\{arg\_role\}' or answer 'N/A' if the question is not answerable. \textbackslash{}nPassage: \{sentence\}.                                                              \\ & Question: What is the '\{arg\_role\}' for the '\{e\_type\}' event? \\  \midrule
\multirow{3}{*}{Layer-2 comparator}                        & Given a news article about '\{e\_type\}' and two candidate spans, decide whether '\{arg1\}' is a more      \\ & acceptable '\{arg\_role\}' than '\{arg2\}' for the '\{e\_type\}' event. \textbackslash{}nArticle: \{article\} \textbackslash{}n  \\  & For this '\{e\_type\}' event, is '\{arg1\}' a more acceptable '\{arg\_role\}' than '\{arg2\}'? Answer yes/no. \\ \midrule
\multirow{2}{*}{Document-level extractor} & Given a news article about \{e\_type\}, select the tokens representing information about '\{arg\_role\}'. \textbackslash{}n      \\ &   Context: \{news\}. Question: What is the '\{arg\_role\}' for the '\{e\_type\}' event?  \\ \bottomrule
\end{tabular}
}
\caption{Instructions designed for each stage in \model. The document-level extractor is utilized in the ensembling mode of \model (\Cref{sec:ensemble}), and serves as the Flan-UL2 baseline (\cref{sec:experiments}). These \textit{aligned instructions} are adapted from task 179 (participant extraction) in NIv2\footnote{https://raw.githubusercontent.com/allenai/natural-instructions/master/tasks/README.md} \citep{wang-etal-2022-super}.}
\label{tbl:prompts}
\end{table*}

\begin{table*}[]
\centering
\resizebox{.95\linewidth}{!}{%
\begin{tabular}{lll} \toprule
Extraction & Ground Truth                                    & Judgements                                                 \\ \midrule
N/A        & N/A                                   & Yes.                                                       \\
something  & N/A                                   & No, you should generate ``N/A''                               \\
N/A        & something                             & No, you should generate ``{[}GT{]}''.                          \\
something  & something                             & Yes.                                                       \\
something  & something longer OR something shorter & You are almost there! The right answer should be ``{[}GT{]}''. \\
anything   & something                             & No, you should generate ``{[}GT{]}''.     \\ \bottomrule                    
\end{tabular}
}
\caption{Designed template-based judgments used to train the \leafer module in order to address the boundary identification issue. We categorize the (extraction, ground truth) pairs into six classes. Here, ``anything'' refers to a generated extraction that is completely off, and ``{[}GT{]}'' acts as a placeholder to be replaced with a specific ground-truth argument.}
\label{tbl:judgements}
\end{table*}



{\section{Cost Estimation}
\label{sec:cost_estimate}
In addition to models' extraction performance, Table~\ref{tbl:results} also presents the cost estimation of each model family. We briefly introduce the criteria used when estimating monetary costs. The training cost is mainly associated with document annotations.\footnote{Here, we omit the sunk cost incurred due to pre-training.} Regarding inference, we consider the expenses incurred in hitting API endpoints.\footnote{The server maintenance cost is considered low ($\sim0$).} Per \citet{tong-etal-2022-docee}, both EEQA and Ontology QA are trained on 22K articles, each costs $\$0.9$, totaling $\$20,000$.
 Based on ChatGPT pricing,\footnote{\url{https://azure.microsoft.com/en-us/pricing/details/cognitive-services/openai-service/}} the base cost is \$0.004/1K tokens. Processing each article and then producing answers would consume 5K to 50K tokens on average, depending on the input mode. The test set contains 2K examples, so the total cost is around $\$40$ to $\$400$. For the Flan-UL2 baseline and our \model, each only needs annotations of up to 50 articles, for the training of the \leafer module. It is noteworthy that \model enables cost-effective scaling at inference, while ChatGPT might face budget constraints. 
}

\label{sec:appendix}

\end{document}